\documentclass[10pt,twocolumn,letterpaper]{article}

\usepackage{cvpr}
\usepackage{times}
\usepackage{epsfig}
\usepackage{graphicx}
\usepackage{amsmath}
\usepackage{amssymb}
\usepackage{booktabs}
\usepackage{pifont}
\usepackage{multirow}
\usepackage[dvipsnames]{xcolor}
\usepackage{verbatim}

\newcommand{\cmmnt}[1]{\ignorespaces}

\newcommand{\cmark}{\ding{51}}%

\usepackage[pagebackref=true,breaklinks=true,colorlinks,bookmarks=false,pdftitle={Video to Events: Recycling Video Datasets for Event Cameras}, pdfsubject={Computer Vision, Event Camera, Generative Model}, pdfauthor={D. Gehrig, M. Gehrig, J. Hidalgo-Carrio, D. Scaramuzza}, pdfkeywords={Computer Vision, Event Cameras, Generative Model}]{hyperref}

\cvprfinalcopy 

\def\cvprPaperID{6997} 

\ifcvprfinal\fi

\title{Video to Events:\\Recycling Video Datasets for Event Cameras}
\definecolor{somegray}{rgb}{0.5, 0.5, 0.5}
\newcommand{\darkgrayed}[1]{\textcolor{somegray}{#1}}
\makeatletter
\newcommand*\titleheader[1]{\gdef\@titleheader{#1}}
\AtBeginDocument{%
  \let\st@red@title\@title
  \def\@title{%
    \vskip-3em
    \bgroup\normalfont\large\centering\@titleheader\par\egroup
    \vskip1.5em\st@red@title}
}
\makeatother

\titleheader{\darkgrayed{This paper has been accepted for publication at the\\
IEEE Conference on Computer Vision and Pattern Recognition (CVPR), Seattle, 2020.
\copyright IEEE}}

\title{Video to Events:\\Recycling Video Datasets for Event Cameras}

\begin{document}

\author{Daniel Gehrig\thanks{Equal contribution}\qquad Mathias Gehrig$^{*}$ \qquad Javier Hidalgo-Carri\'o \qquad Davide Scaramuzza\\
Dept. Informatics, Univ. of Zurich and \\
Dept. of Neuroinformatics, Univ. of Zurich and ETH Zurich\\
}

\maketitle
\thispagestyle{empty}

\begin{abstract}
Event cameras are novel sensors that output brightness changes in the form of a stream of asynchronous ``events" instead of intensity frames. They offer significant advantages with respect to conventional cameras: high dynamic
range (HDR), high temporal resolution, and no motion blur. 
Recently, novel learning approaches operating on event data have achieved impressive results. Yet, these methods require a large amount of event data for training, which is hardly available due the novelty of event sensors in computer vision research.
In this paper, we present a method that addresses these needs by converting any existing video dataset recorded with conventional cameras to \emph{synthetic} event data. This unlocks the use of a virtually unlimited number of existing video datasets for training networks designed for real event data. 
We evaluate our method on two relevant vision tasks, i.e., object recognition and semantic segmentation, and show that models trained on synthetic events have several benefits: 
(i) they generalize well to real event data, even in scenarios where standard-camera images are blurry or overexposed, by inheriting the outstanding properties of event cameras;
(ii) they can be used for fine-tuning on real data to improve over state-of-the-art for both classification and semantic segmentation\cmmnt{by 4.3\% and 0.8\%}. 
\end{abstract}
\section*{Multimedia Material}
This project's code is available at\\\url{https://github.com/uzh-rpg/rpg_vid2e}. Additionally, qualitative results are available in this video: \url{https://youtu.be/uX6XknBGg0w}

\section{Introduction}

Event cameras, such as the Dynamic Vision Sensor~\cite{Lichtsteiner08ssc} (DVS), are novel sensors that work radically differently from
conventional cameras. Instead of capturing intensity images at a fixed rate, they measure \emph{changes} of intensity
asynchronously at the time they occur. This results in a stream of \emph{events}, which encode the time, location, and polarity (sign) of brightness changes.
%
\begin{figure}[t]
    \centering
	\includegraphics[width=\linewidth]{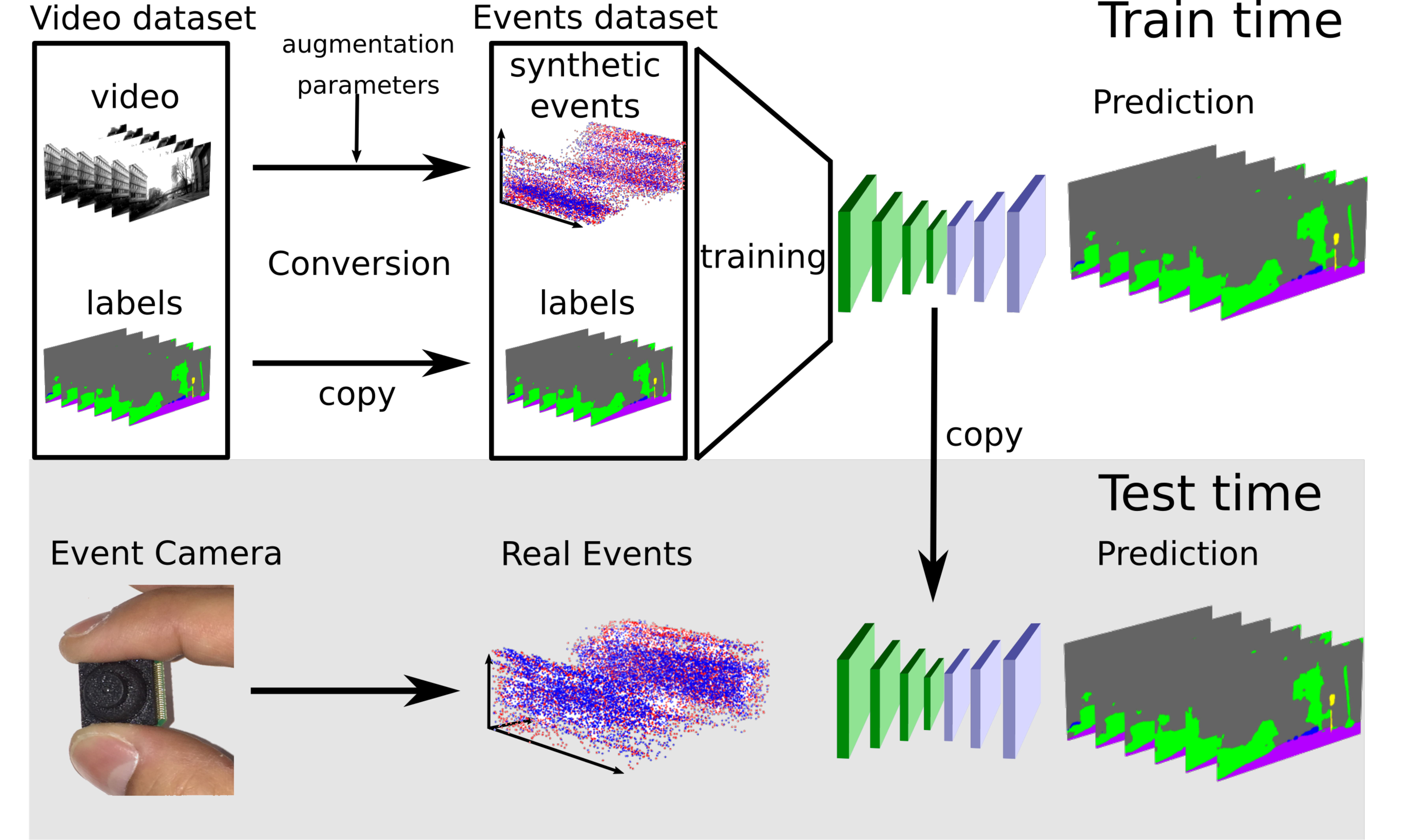}
    \caption{Our method converts any large scale, high quality video dataset, to a synthetic event camera dataset. This unlocks the great wealth of existing 
    video datasets for event cameras, enabling new and exciting applications, and addressing the shortage of high quality event camera datasets.
    Networks trained on these synthetic events generalize surprisingly well to real events. By leveraging the high dynamic range
    and lack of motion blur of event cameras these networks can generalize to situations where standard video frames over exposed or blurred. Best viewed in color.}
    \label{fig:eyecatcher}
    \vspace{-2ex}
\end{figure}

Event cameras possess outstanding properties when compared
to conventional cameras. They have a very high dynamic
range (140 dB versus 60 dB), do not suffer from motion blur, and provide measurements with a latency on the order of microseconds. Thus, they are a viable alternative, or complementary sensor, in conditions that are challenging for standard cameras, such as high-speed and high-dynamic-range (HDR) scenarios~\cite{Kim14bmvc,Rosinol18ral,Zhu19cvpr,Rebecq19cvpr}.\footnote{\url{https://youtu.be/0hDGFFJQfmA}} 
However, because the output of event cameras is asynchronous, existing computer vision algorithms developed for standard cameras cannot be directly applied to these data but need to be specifically tailored to leverage event data (for a survey on event cameras and the field of event-based vision, we refer the reader to~\cite{Gallego19arxiv}).
 
Recently, novel learning approaches operating on event data have achieved impressive results in scenarios where networks operating on standard cameras fail~\cite{Maqueda18cvpr,Zhu18rss,Alonso19cvprw,Zhu19cvpr,Rebecq19cvpr,Gehrig19iccv,Rebecq19arxiv}. 
Notably, in~\cite{Rebecq19arxiv} it was shown that a network trained to reconstruct grayscale intensity frames solely from events can synthesize high framerate videos ($>5,000$ frames per second) of high-speed phenomena (e.g., a bullet shot by gun hitting an object) and can as well render HDR video in challenging lighting conditions (e.g., abrupt transition from dark to bright scene). 
It was also shown that off-the-shelf deep learning algorithms trained on large-scale standard camera datasets can be applied to these synthesized HDR, high-framerate videos and that, by doing so, they consistently outperforms algorithms that were specifically trained only on event data.\footnote{\url{https://youtu.be/eomALySSGVU}}
These results highlight that the \emph{event data contain all the visual information} that is needed to carry out the same tasks that can be accomplished with standard cameras and that it should be possible to design \emph{efficient learning algorithms} that process the event data \emph{end to end without passing through intermediate image representations}. 

Unfortunately, the design of efficient, end-to-end learning methods requires a large amount of event data for training, which is hardly available because of the novelty of event sensors: event cameras were first commercialized in 2008 and research on event-based vision has made most progress only in the past five years. 
A viable alternative to the lack of large scale datasets are event camera simulators~\cite{Rebecq18corl}; however, an open research question is how well neural networks trained on synthetic events will generalize to real event cameras. Moreover, simulated scenarios still suffer from lack of realism.

To address these issues, we propose a method to generate synthetic, large-scale event-camera data from existing real-world, video datasets recorded with conventional cameras.
On the one hand, our method addresses the shortage of event-camera data by leveraging the virtually unlimited supply of existing video datasets and \emph{democratizing}  this data  for event camera research. The availability of these new datasets can unlock new and exciting research directions for event cameras and spark further research in new fields, previously inaccessible for event cameras. 
On the other hand, since our method directly relies on video sequences recorded in real-world environments, 
we show that models trained on synthetic events generated from video generalize surprisingly well to real event data, even in challenging scenarios, such as HDR scenes or during 
fast motions. 
To conclude, our contributions are:

\begin{itemize}
    \item We present a framework for converting existing video datasets to event datasets, thus enabling new applications for event cameras.
    \item We show that models trained on these synthesized event datasets generalize well to real data, even in scenarios where standard images are blurry or overexposed, by inheriting the outstanding properties of event cameras.
    \item We evaluate our method on two relevant vision tasks, i.e., object recognition and semantic segmentation, and show that models trained on synthetic events can be used for fine-tuning on real data to improve over state of the art.
\end{itemize}

Our work is structured as follows: First, we review relevant literature in event camera research and deep learning techniques 
as well as available datasets in Sec. \ref{sec:related_work}. We then present the method for converting video datasets to events in Sec. \ref{sec:method}.
Section \ref{sec:classification} validates and characterizes the realism of events generated by our approach in the setting 
of object recognition (Sec. \ref{sec:classification}). Finally, we apply our method to the challenging task of per-pixel semantic segmentation in
Sec. \ref{sec:semantic_segmentation}.

\section{Related Work}\label{sec:related_work}
\subsection{Event Camera Datasets for Machine Learning}
The number of event camera datasets tailored to benchmarking of machine learning algorithms is limited. The earliest such datasets are concerned with \emph{classification} and are counterparts of their corresponding image-based datasets. Both Neuromorphic (N)-MNIST and N-Caltech101 \cite{Orchard15fns} were generated by mounting an event camera on a pan-and-til unit in front of a monitor to reproduce the saccades for generating events from a static image. Later, Sironi et al. \cite{Sironi18cvpr} introduced N-CARS, a binary classification dataset but with events from dynamic scenes rather than static images. The most recent classification dataset \cite{Bi19iccv}, termed American Sign Language (ASL)-DVS, features 24 handshapes for american sign language classification. Closely related to neuromorphic classification is neuromorphic action recognition. This task has been targeted by the DVS-Gesture dataset \cite{Amir17cvpr} which contains 11 different gestures recorded by the DVS128 event camera.\\
The first and so far only neuromorphic human pose dataset, DAVIS Human Pose Dataset(DHP19), has been recently introduced by \cite{calabrese2019dhp19}. It features four event cameras with resolution of $260\times346$ recording 33 different movements simultaneously from different viewpoints.\\
The DAVIS Driving Dataset (DDD17) \cite{Binas17icml} and Multi-Vehicle Stereo Event Camera (MVSEC) dataset \cite{Zhu18ral} are two driving datasets. The former provides data about vehicle speed, position, steering angle, throttle and brake besides a single event camera. The latter dataset features multiple vehicles in different environments and also provides ego-motion and LIDAR data together with frames and events from a stereo DAVIS setup. A subset of DDD17 was later extended \cite{Alonso19cvprw} with approximate semantic labels to investigate semantic segmentation with event cameras.

\subsection{Deep Learning with Event Cameras}

The applicability of deep learning to event camera data was first explored in the context of classification. Neil et al. \cite{Neil16nips} designed a novel recurrent neural network architecture applied to classification on the N-MNIST dataset. Later, Maqueda et al. \cite{Maqueda18cvpr} proposed an event-frame representation and designed a CNN architecture for steering angle regression on the DDD17 dataset. The same dataset has been modified by Alonso et al. \cite{Alonso19cvprw} to perform semantic segmentation. The availability of MVSEC has spurred research in optical flow \cite{Zhu18rss, Zhu19cvpr, Gehrig19iccv} and depth estimation \cite{Zhu19cvpr, tulyakov2019}. In contrast to aforementioned work, \cite{Rebecq19cvpr, Rebecq19arxiv} trained a convolutional recurrent neural network entirely on simulated events to perform image reconstruction.

\subsection{Synthetic Events}
This section reviews work in the domain of generative modeling for events from event cameras. Early work in this domain has been performed by Kaiser et al. \cite{Kaiser16simpar}. They generate events simply by applying a threshold on the image difference. Depending on the pixel's intensity difference a positive or negative event is generated. Pix2NVS~\cite{Bi17icip} computes per-pixel luminance from conventional video frames. The technique generates synthetic events with inaccurate timestamps clustered to frame timestamps. To the best of our knowledge, the two first simulators attempting to generate events accurately are \cite{Mueggler17ijrr} and \cite{Li18bmvc}. Both works render images at high frame-rate and linearly interpolate the intensity signals to generate events. Rebecq et al. \cite{Rebecq18corl} additionally introduces an adaptive sampling scheme based on the maximum displacement between frames. This leads to improved accuracy for very fast motion and lower computation in case of slow motion. The generative model used in \cite{Mueggler17ijrr, Rebecq18corl} has been formalized in previous work \cite{Lichtsteiner08ssc, Gallego15arxiv, Gallego17pami}.
\section{Methodology}
\label{sec:method}
\begin{figure*}[t]
    \centering
    \includegraphics[width=0.7\linewidth]{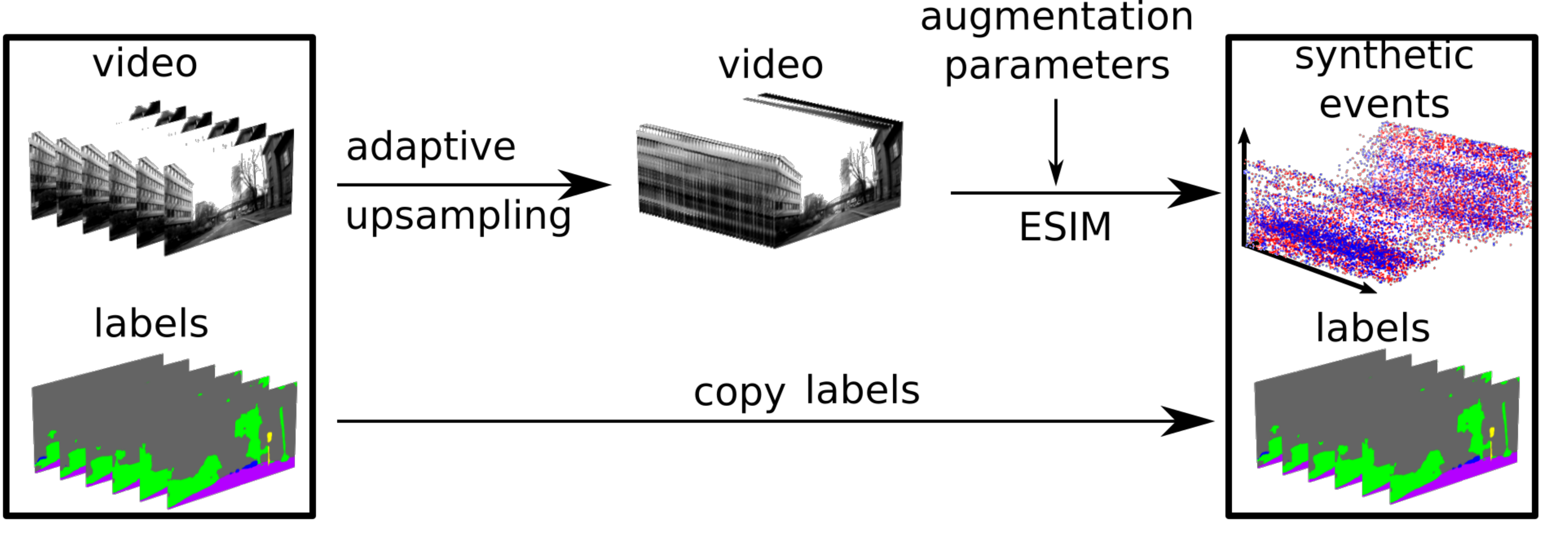}
    \caption{Overview of the method. Low frame-rate video is first adaptively upsampled using 
    the method proposed in \cite{Jiang18cvpr}. This upsampled video is fed to the event camera simulator (ESIM) 
    \cite{Rebecq18corl} which produces asynchronous and sparse events with high temporal resolution.}
    \label{fig:method_general}
    \vspace{-2ex}
\end{figure*}

In this section, we describe our method for converting video to synthetic events.
This conversion can be split into two steps: event generation and frame upsampling, covered in Sec. \ref{sec:event_generation} and Sec. \ref{sec:frame_upsampling}, respectively. Fig. \ref{fig:method_general} illustrates
these individual steps. In a first step, we leverage a recent frame interpolation technique \cite{Jiang18cvpr} to convert low frame rate to high frame rate video using an adaptive upsampling technique. This video is then used to generate events using the generative model by leveraging a recent event camera simulator (ESIM) \cite{Rebecq18corl}. To facilitate domain adaptation between synthetic and real events, we further introduce two domain adaptation techniques. Finally, we make use of \cite{Gehrig19iccv} to convert the sparse and asynchronous events to tensor-like representations, which enables learning with traditional convolutional neural network (CNN) architectures.
%
%
\subsection{Event Generation Model}
\label{sec:event_generation}
Event cameras have pixels that are independent and respond to changes in the
continuous log brightness signal $L(\textbf{u}, t)$. An event $e_k=(x_k, y_k, t_k, p_k)$ is triggered when the magnitude of the log brightness at pixel
$u=(x_k, y_k)^T$ and time $t_k$ has changed by more than a threshold $C$ since
the last event at the same pixel. 
\begin{equation}
	\label{eq:generative_model}
	\Delta L(\textbf{u}, t_k) = L(\textbf{u}, t_k)-L(\textbf{u}, t_k-\Delta t_k) \geq p_k C.
\end{equation}
Here, $\Delta t_k$ is the time since the last triggered event, $p_k\in\{-1,+1\}$
is the sign of the change, also called polarity of the event. Equation
\eqref{eq:generative_model} describes the generative event model for an ideal
sensor \cite{Gallego17pami, Gallego19arxiv}. 

\subsection{Frame Upsampling}
\label{sec:frame_upsampling}
While the event generative model provides a tool for generating events
for a given brightness signal, it requires that this signal be known at high
temporal resolution. In particular for event cameras this timescale is on the order of
microseconds. Event camera simulators, such as ESIM, can address this problem
by adaptively rendering virtual scenes at arbitrary temporal resolution
(Section 3.1 of \cite{Rebecq18corl}). However, video sequences typically only provide 
intensity measurements at fixed and low temporal resolution on the order of milliseconds. 

We therefore seek to recover the full intensity profile $I(\textbf{u}, t)$ given
a video sequence of $N$ frames $\{I(\textbf{u}, t_i)\}_{i=0}^{N}$ captured at
times $\{t_i\}_{i=0}^{N}$. A subproblem using only two consecutive frames has been well 
studied in frame interpolation literature. We thus turn to \cite{Jiang18cvpr}, a recent 
technique for frame interpolation which is finding wide spread use in smart-phones. Compared 
to other frame interpolation techniques such as \cite{Liu17iccv, Long16eccv, Niklaus17cvpr, Niklaus17iccv}
the method in \cite{Jiang18cvpr} allows to reconstruct frames
at arbitrary temporal resolution, which is ideal for the posed task.
The number of intermediate frames, must be chosen carefully since too low values lead to aliasing of the brightness signal (illustrated in \cite{Rebecq18corl}, Fig. 3) but too high values impose a computational burden.
The following adaptive sampling strategy, inspired by \cite{Rebecq18corl}, uses bidirectional optical flow (as estimated internally by \cite{Jiang18cvpr}) to compute the number of intermediate samples. Given two consecutive frames $I(t_i)$ and $I(t_{i+1})$ at times $t_i$ and $t_{i+1}$, we generate $K_i$ equally spaced intermediate frames. $K_i$ is chosen such that the relative displacement between intermediate frames is at most 1 pixel for all pixels:
\begin{equation}
\label{eq:interpolation}
K_i = \max_{\textbf{u}}{\max\{\Vert\mathbf{F}_{i\rightarrow i+1}(\mathbf{u})\Vert,\Vert\mathbf{F}_{i+1\rightarrow i}(\mathbf{u})\Vert\}} - 1,
\end{equation}
where $\mathbf{F}_{i\rightarrow j}(\mathbf{u})$ is the optical flow from frame $i$ to $j$ at pixel location $\mathbf{u}$.
We use this strategy to adaptively upsample between pairs of video frames, resulting in an adaptively upsampled video sequence (Fig. \ref{fig:method_general} middle).

\subsection{Event Generation from High Frame Rate Video}
\label{sec:event_generation_from_frames}
The next step is to generate events from the high frame rate video sequence generated in Sec. \ref{sec:frame_upsampling}.
We generate events by employing the algorithm described in \cite{Rebecq18corl} (Sec 3.1). For each pixel the continuous 
intensity signal in time is approximated by linearly interpolating between video frames. Events are generated at each pixel whenever the magnitude of the change in intensity exceeds the contrast threshold, $C$ (defined in \eqref{eq:generative_model}) which is a parameter of ESIM. Since the contrast threshold in \eqref{eq:generative_model} is typically not known for real sensors and can vary from sensor to sensor and between positive and negative events, we propose to randomize it at train-time. Before generating a sequence of events we randomly sample contrast thresholds for positive and negative events, $C_p, C_n$ from the uniform distribution $\sim \mathcal{U}(C_{\text{min}}, C_{\text{max}})$. A similar procedure was used in \cite{Rebecq19cvpr,Rebecq19arxiv} where randomization was shown to improve
domain adaptation between simulated and real data. In this work we chose $C_\text{min}=$0.05 and $C_\text{max}=$0.5.

\subsection{Event Representation and Learning}
As a next step, The synthetic events and original labels are used to train a network. To do this, we consider a window of events leading up to the time stamped ground truth label and train a model to predict it. Note that this works for general datasets with precisely timestamped images and labels.
We take advantage of existing CNN architectures designed for standard images by converting the asynchronous and sparse event streams into tensor-like representation. We chose the Event Spike Tensor (EST) \cite{Gehrig19iccv} since it was shown to outperform existing representations on both high- and low-level tasks. The EST is generated by drawing the events with positive and negative polarity into two separate spatio-temporal grids of dimensions $H\times W\times C$ and stacking them along the channel dimension. Here $H$ and $W$ are the sensor resolution and $C$ is a hyper-parameter which controls the number of temporal bins used to aggregate events. In this work we chose $C=15$.

\vspace{-1ex}
\section{Experiments}
In this section, we present an evaluation of the method described in \ref{sec:method} on two tasks: object classification (Sec. \ref{sec:classification}) and semantic segmentation (Sec. \ref{sec:semantic_segmentation}). In each case we show that models that are trained on synthetic events have the following benefits: (i) they generalize well from synthetic to real events (ii), can be used to fine tune on real event data, leading to accelerated learning and improvements over the state of the art, and (iii) can generalize to scenarios where standard frames are blurry or underexposed.
\subsection{Object Recognition}
\label{sec:classification}
Object recognition using standard frame-based cameras remains challenging due to their
low dynamic range, high latency and motion blur. Recently, event-based object
recognition has grown in popularity since event cameras address all of these challenges.
In this section we evaluate the event generation method proposed in
\ref{sec:method} in this scenario.
In particular, we provide an analysis of each component of the method, frame upsampling and event generation. In our evaluation we use N-Caltech101 (Neuromorphic-Caltech101) \cite{Orchard15fns}, the event-based version of the popular Caltech101 dataset \cite{Li06pami} which poses the task of multi class recognition.
This dataset remains challenging due to a large class imbalance.
The dataset comprises 8,709 event sequences from from 101 object classes each lasting for the duration of 300 ms. 
Samples from N-Caltech101 were recorded by placing an event camera in front of a screen and projecting various examples from Caltech101, while the event camera underwent three saccadic movements. 
\vspace{-1ex}
\subsubsection{Implementation}
To evaluate our method we convert the samples of Caltech101 to event streams, thus generating a replica (sim-N-Caltech101) of the N-Caltech101 dataset. We then aim at quantifying how well a network trained on sim-N-Caltech101 generalizes to events in the real dataset, N-Caltech101. To convert samples from Caltech101 to event streams we adopt the strategy for converting still images to video sequences outlined in \cite{Rebecq19cvpr, Rebecq19arxiv}. We map the still images onto a 2D plane and simulate an event camera moving in front of this plane in a saccadic motion, as was done for the original N-Caltech101 dataset \cite{Orchard15fns}. Note that, since the camera is moved virtually, video frames can be rendered at arbitrary temporal resolution, making it possible to simulate video cameras with different frame rates. Once a high frame rate video is rendered, we use this video to generate events. In a first step we fix the contrast threshold in ESIM to 0.06 but randomize this value later. Some examples from sim-Caltech101 as well as corresponding samples from N-Caltech101 and Caltech101 are shown in Fig. \ref{fig:sim_ncaltech_examples}. 

In a next step we train a classifier on data from sim-N-Caltech101. We chose an off-the-shelf classifier based on ResNet-34 \cite{He16cvpr} which has been pretrained on RGB images from ImageNet \cite{Russakovsky15ijcv}. We choose a batch size of 4 and a learning rate of $10^-6$ and trained the network to convergence. We then compute the test score of this network on a held out set on the real dataset which is reported in the first row of Tab. \ref{tab:ncaltech_rand}. As a baseline we compare against a network which was trained on real data and evaluated on the same held out test set. We can observe that the network trained on  synthetic events leads to a lower score ($75.1\%$) than one trained on real events ($86.3\%$) leading to a gap of $11.2\%$. 
To address this gap we apply a form of domain randomization by randomly sampling the contrast threshold during training, as was described in \ref{sec:event_generation_from_frames}. This is done for two reasons: On the one hand this step helps to add robustness to the network by exposing it to a larger variety of event streams, which benefits generalizability. On the other hand the true contrast threshold it typically not known during training, so randomization eliminates the need for hand tuning this parameter. By employing this technique we achieve an improved result of $78.2\%$ reducing the gap to $8.1\%$. 

We propose to further generalizability through dataset extension. It is well known that 
Caltech101 is unbalanced. For example, while the most common class (airplanes) has 800 samples, the least 
common class (inline skate) has only 31 samples. To address this imbalance, we exploit the fact that our method
does not require real events. We downloaded images from the Internet (google images) to find additional examples for each class. We filtered wrong samples by using a ResNet-34 classifier \cite{He16cvpr}, pretrained on Caltech101 images. By employing this strategy without contrast threshold randomization we achieve a test score of $76.9\%$  and if we include both techniques we achieve a score of $80.7\%$, effectively reducing the gap to real to $5.6\%$. While this gap still remains, this result shows that the synthetic events generated by our method effectively capture most of the visual appearance of the real event stream thus achieve a high level of realism. 
\vspace{-1ex}
\paragraph{Fine Tuning}
In this section we show that a network pretrained on simulated data described in the previous section can be used to fine tune on real data, which leads to a large performance increase. We fine tune the best model obtained in the previous experiment network obtained by training on real events from N-Caltech101 with a reduced learning rate of $10^{-7}$ and train until convergence. The test score is reported in Tab. \ref{tab:ncaltech_rand} where we see that fine tuning has a large impact on network performance. Not only does the test score surpass baseline on real data, it also beats existing state of the art event-based approaches, summarized in Tab. \ref{tab:ncaltech_comp}, such as \cite{Sironi18cvpr, Gehrig19iccv, Rebecq19cvpr} and approaches state of the art methods using standard images \cite{Mahmood17icip} with 94.7\%.
\vspace{-1ex}
\subsubsection{Effect of Frame Upsampling}
\label{sec:exp_upsampling}
\begin{figure}[t]
    \centering
    \begin{tabular}{c}
        \includegraphics[width=\linewidth]{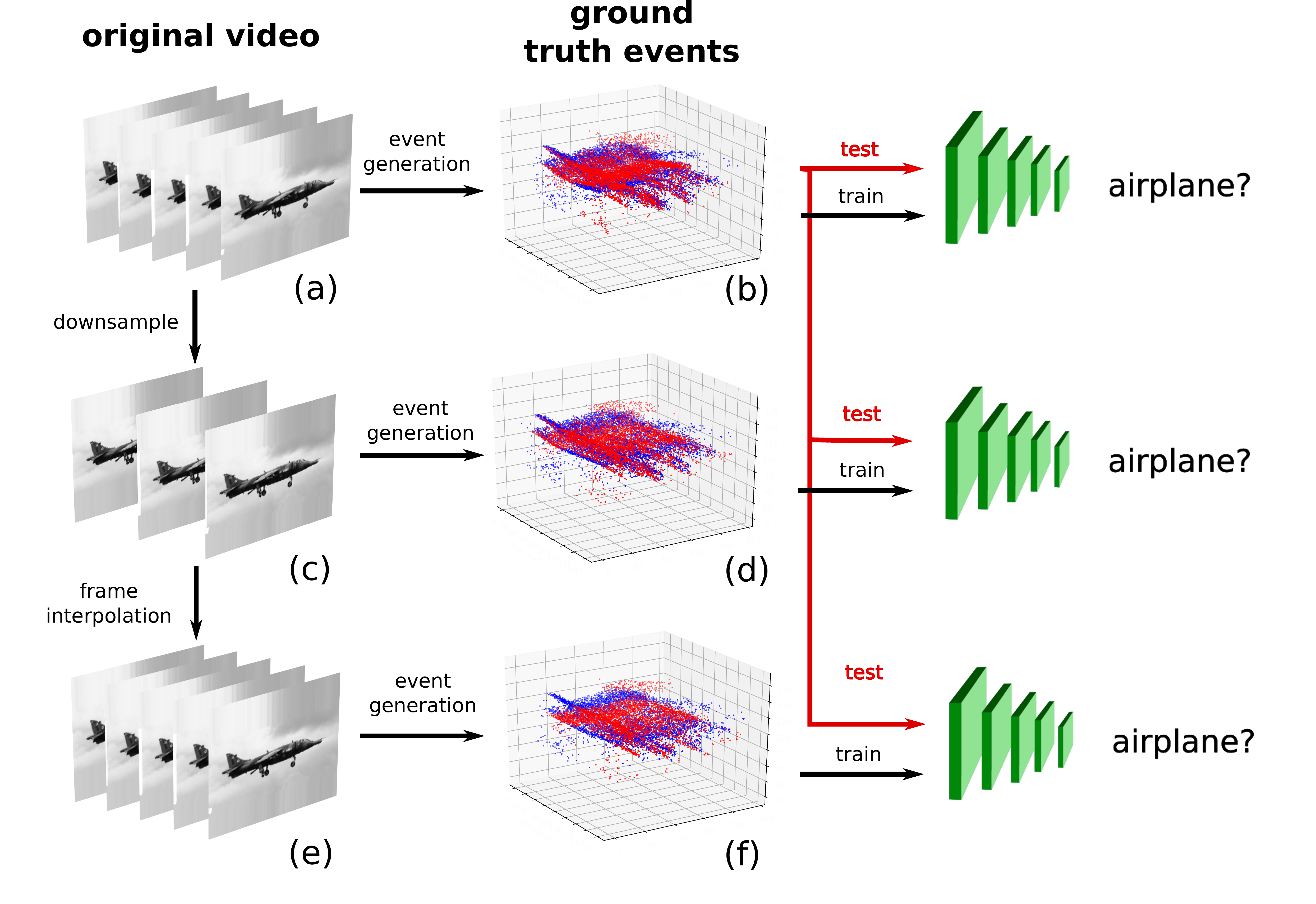}
    \end{tabular}
    \caption{Evaluation of the effect of frame interpolation on event quality.
    We render a high frame rate video of a Caltech101 \cite{Li06pami} image  (a) by sliding a virtual camera
    in front of a 2d planes, following three saccadic movements as described in \cite{Orchard15fns} 
    which we use to generate ground truth events (b). We then downsample the video (c) which
    leads to a distortion of the event stream (d). By applying the interpolation technique
    in \cite{Jiang18cvpr} we can reconstruct the original video (e) which leads to improved
    event quality. To quantify this quality, we train three classifiers, one on each dataset, and compare 
    test scores on the ground truth events.}
    \label{fig:updownsampling}
\end{figure}

\begin{figure}[t]
    \centering
    \begin{tabular}{ccc}
        \includegraphics[width=0.3\linewidth]{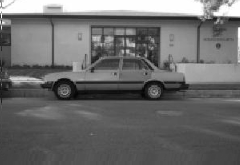} &
        \includegraphics[width=0.3\linewidth]{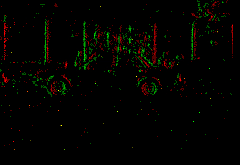}&
        \includegraphics[width=0.3\linewidth]{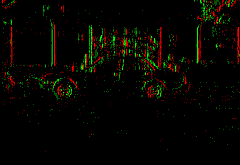}\\
        \includegraphics[width=0.3\linewidth]{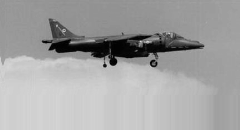}&
        \includegraphics[width=0.3\linewidth]{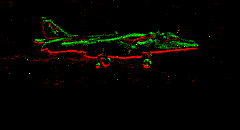}&
        \includegraphics[width=0.3\linewidth]{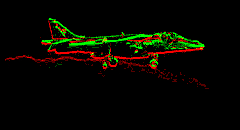}\\
        \includegraphics[width=0.3\linewidth]{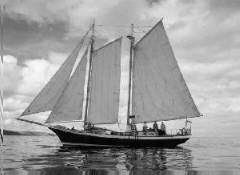}&
        \includegraphics[width=0.3\linewidth]{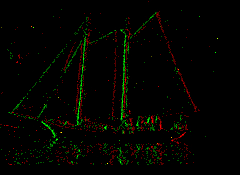}&
        \includegraphics[width=0.3\linewidth]{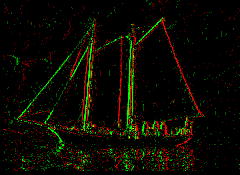}\\
        
        (a) preview & (b) real events & (c) synthetic events
    \end{tabular}
    \caption{A side-by-side comparison of a samples from Caltech101 (a),
    N-Caltech101 (b) and our synthetic examples from (c) sim-N-Caltech101. While
    the real events were recorded by moving an event camera in front of a
    projector, the synthetic events were generated using ESIM by moving a virtual camera in front of a 2D projection of the
    sample in (a). }
    \label{fig:sim_ncaltech_examples}
\end{figure}

In this section we present an ablation study which aims at characterizing the effect of frame upsampling on the generated events. This is crucial since our method relies on video sequences, which typically only record visual information at low temporal resolution. In particular, we show that adaptive frame upsampling leads to improvements in the events in the case of low frame rate video. To understand this relationship we propose the following controlled experiment, illustrated in Fig.~\ref{fig:updownsampling}. We first generate a reference dataset from Caltech101 samples by rendering video frames at 530 Hz (Fig. \ref{fig:updownsampling} a), for 300 ms, such that the maximal displacement between consecutive frames is below 1 pixel (0.13 pixel). We simulate the low frame rate of conventional video cameras by downsampling these frames by factors of (4, 16, and 80) leading to maximal pixel displacements of 0.55, 2.11 and 9.4 respectively (Fig. \ref{fig:updownsampling} c). To recover high frame rate video we apply the frame interpolation technique described in \cite{Jiang18cvpr}, which results in frames at the same temporal resolution as the original video. To understand the effect of video quality on events we generate datasets for each of these three cases, fixing the settings
for event generation, and varying the downsampling factor. This way changes in the events are reflected by the changes in video quality. To assess these differences we train three classifiers with the same training and network parameters as described in the previous section, and compare their test scores on events generated from the original high frame rate video. The test scores for different downsampling factors is reported in Tab. \ref{tab:ncaltech_upsampling}. While a network trained on events from high frame rate video (Tab.~ \ref{tab:ncaltech_upsampling} top row) achieves a high score of $88.6\%$ on this test set, we see that reducing the framerate (Tab.~ \ref{tab:ncaltech_upsampling} second row) by a factor of 80, drastically reduces this score to $61.8\%$. In fact, artifacts caused by the low frame rate become apparent at these low frame rates. One such artifact is called ghosting and is caused when there is a large displacement between consecutive frames. In this case linear interpolation of the intensity values over time results in the appearance and disappearance of parts of the scene which cause events to be generated in an unrealistic fashion. By using frame interpolation we reduce these effects, as indicated by the increased performance ($68.7\%$).\\
%
%
\begin{table}[]
\centering
  \resizebox{0.4\textwidth}{!}{  
    \begin{tabular}{c|cccc}
        \toprule
        \multirow{2}{*}{\textbf{\begin{tabular}[c]{@{}c@{}}video\end{tabular}}}         & \multicolumn{4}{c}{\textbf{downsampling factor}}                                                         \\ \cline{2-5} 
                                                                                                                 & 1                        & 4                        & 16                       & 80                      \\ \midrule
        original                                                                                                 & 0.887                    & \multicolumn{3}{c}{-}                                                         \\
        downsampled                                                                                              & 0.887                    & 0.882                    & 0.867                    & 0.618                   \\
        interpolated                                                                                             & 0.887                    & 0.881                    & 0.877                    & 0.687                   \\ \midrule
        \multicolumn{1}{l|}{\begin{tabular}[c]{@{}l@{}}average interframe \\ displacement {[}px{]}\end{tabular}} & \multicolumn{1}{l}{0.13} & \multicolumn{1}{l}{0.55} & \multicolumn{1}{l}{2.11} & \multicolumn{1}{l}{9.4} \\ \bottomrule
    \end{tabular}
    }
    \caption{Ablation study on the effect of downsampling. Test score of networks trained on events generated from different video streams and evaluated on events from high frame rate video.}
    \label{tab:ncaltech_upsampling}
    \vspace{-2ex}
\end{table}
%

%
\begin{table}[]
\centering
\resizebox{0.4\textwidth}{!}{  
    \begin{tabular}{ccc|c}
        \toprule
        \textbf{\begin{tabular}[c]{@{}c@{}}contrast threshold \\ randomization\end{tabular}} & \textbf{\begin{tabular}[c]{@{}c@{}}dataset \\ extension\end{tabular}} & \textbf{\begin{tabular}[c]{@{}c@{}}fine tuning \\ on real\end{tabular}} & \textbf{test score} \\ \midrule
              &      &      &0.751 \\
              &\cmark&      &0.769 \\
        \cmark&      &      &0.782 \\
        \cmark&\cmark&      &0.807 \\
              &      &\cmark&0.856 \\
              &\cmark&\cmark&0.852 \\
        \cmark&      &\cmark&0.904 \\
        \cmark&\cmark&\cmark&\textbf{0.906}\\ \midrule
        \multicolumn{3}{c|}{real data}                                                                                                                                                                                                        & 0.863               \\ 
        \multicolumn{3}{c|}{images \cite{Mahmood17icip}}                                                                                                                                                                                                           & 0.947               \\ \bottomrule
    \end{tabular}
}
\caption{Effect of randomization on test accuracy. For comparison we 
report the test scores when trained on real events and also the state of the art \cite{Mahmood17icip} on the original Caltech101 images.}
\label{tab:ncaltech_rand}

\end{table}
%
\begin{table}[]
\centering
\resizebox{0.4\textwidth}{!}{  
    \begin{tabular}{c|c|c}
        \toprule
        \textbf{Method} & \textbf{Training Data} & \textbf{Test Score} \\ \midrule
        HATS \cite{Sironi18cvpr}            & real & 0.642               \\
        HATS+ResNet-34 \cite{Sironi18cvpr}  & real & 0.691               \\
        RG-CNN \cite{Bi19iccv}              & real & 0.657               \\
        EST \cite{Gehrig19iccv}             & real & 0.817               \\    
        E2VID \cite{Rebecq19cvpr}           & real & 0.866               \\
        \midrule
        ours& synthetic        & 0.807               \\
        ours& synthetic + real & \textbf{0.906}      \\ \bottomrule
    \end{tabular}
}
\caption{Comparison of classification accuracy for state of the art classification methods on N-Caltech101 \cite{Orchard15fns}. Our method uses a ResNet-34\cite{He16cvpr} architecture.}
\label{tab:ncaltech_comp}
\vspace{-5ex}
\end{table}
%
\vspace{-2ex}
\subsection{Semantic Segmentation}\label{sec:semantic_segmentation}

Semantic segmentation is a recognition task which aims at assigning a semantic label to each pixel in an image. It has numerous applications, including street lane and 
pedestrian detection for autonomous driving. Nonetheless, semantic segmentation using standard images remains challenging especially in edge-case scenarios, where their quality is greatly reduced due to motion blur or over- and under-exposure. Event-based segmentation promises to address these issues by leveraging the high dynamic range, lack of motion blur and low latency of the event camera.\\ 

In this section we evaluate our method for semantic segmentation by generating a large scale synthetic event dataset from the publicly available DAVIS Driving Dataset (DDD17) \cite{Binas17icml}. It features grayscale video with events from the Dynamic and Activate Vision Sensor (DAVIS) \cite{Brandli14ssc} and semantic annotations provided by  \cite{Alonso19cvprw} for a selection of sequences. In \cite{Alonso19cvprw} a network trained on Cityscapes \cite{cordts2016cityscapes} was used to generate labels for a total of 19840 grayscale images (15950 for training and 3890 for testing).
The combination of grayscale video and events allows us to generate synthetic events and evaluate against real events from the event camera. We show that training solely on synthetic events generated by our method yields competitive performance with respect the state of the art trained on real events. Furthermore, we improve on the state of the art \cite{Alonso19cvprw} by training on synthetic events and fine tuning on real events. 
\begin{figure*}[ht]
    \centering
    \begin{tabular}{cccc}
        \includegraphics[width=0.2\linewidth]{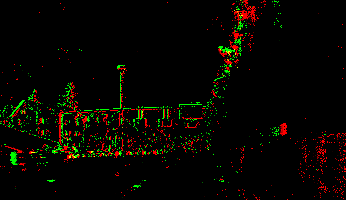}&
        \includegraphics[width=0.2\linewidth]{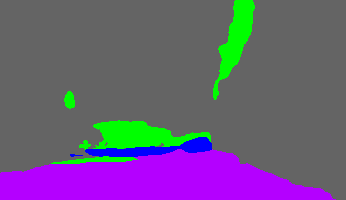}&
        \includegraphics[width=0.2\linewidth]{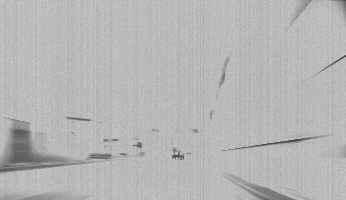}&
        \includegraphics[width=0.2\linewidth]{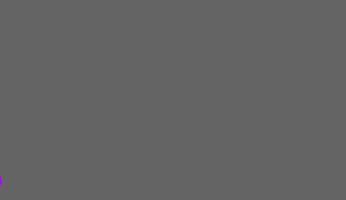}\\
        \includegraphics[width=0.2\linewidth]{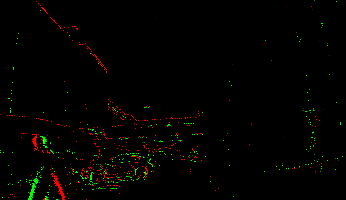}&
        \includegraphics[width=0.2\linewidth]{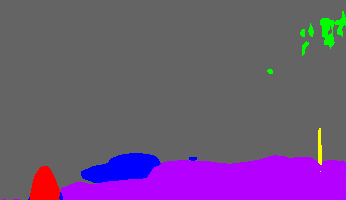}&
        \includegraphics[width=0.2\linewidth]{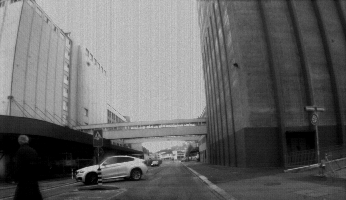}&
        \includegraphics[width=0.2\linewidth]{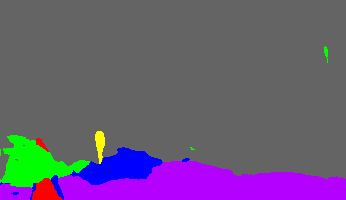}\\
        (a) events & (b) prediction from events & (c) DAVIS frame & (d) prediction from frame
    \end{tabular}
    \caption{Edge cases for semantic segmentation (\textcolor{black}{violet: street}; \textcolor{black}{green: vegetation}; \textcolor{black}{red: person}; \textcolor{black}{blue: car}; \textcolor{black}{yellow: object}; \textcolor{black}{gray: background}). The first row depicts a scenario in which the conventional camera is over-exposed. This results in deteriorated frame-based segmentation performance. In contrast, the event-based segmentation network is able to predict the road labels accurately. The second row showcases a scenario in which  frame-based segmentation wrongly classifies a person as vegetation. This is due to the low contrast in the lower left part of the image. The event camera gracefully handles this case thanks to its superior contrast sensitivity. Best viewed in color.}
    \label{fig:seg_edge_cases}
\end{figure*}
\vspace{-1ex}
\subsubsection{Implementation}
The annotated version of DDD17 \cite{Alonso19cvprw} provides segmentation labels which are synchronized with the frames and thus appear at 10-30 Hz intervals. For each label we use the events that occurred in a 50 ms time window before the label for prediction, as was done in \cite{Alonso19cvprw}. We consider two event-based input representations: the EST, which was already used in Sec. \ref{sec:classification} and the 6-channel representation proposed by \cite{Alonso19cvprw}. In \cite{Alonso19cvprw} a six channel tensor is constructed from the events, with three channels for both the positive and negative events. The first channel is simply the histogram of events; that is the number of events received at each pixel within a certain time interval. The second channel is the mean timestamp of the events while the third channel is the standard deviation of the timestamps. 

We use the network architecture proposed in  \cite{Alonso19cvprw} which consists of a U-Net architecture \cite{Ronneberger15icmicci} with an Xception encoder \cite{chollet2017xception} and a light decoder architecture. We use a batchsize of 8 and use ADAM \cite{Kingma15iclr} with a learning rate of $10^-3$ and train until convergence.\\

\vspace{-4ex}
\subsubsection{Quantitative Results}
As was done in \cite{Alonso19cvprw}, we use the following two evaluation metrics: \emph{Accuracy} and \emph{Mean Intersection over Union (MIoU)}. Given predicted semantic labels $\hat{y}$, ground-truth labels $y$, $N$ the number of pixels and $C$ the number of classes, \emph{accuracy} is defined as
\vspace{-1ex}
\begin{equation}
    \text{Accuracy} = \frac{1}{N}\sum_{n=1}^N\delta(y_n, \hat{y}_n)
\end{equation}
and simply measures the overall proportion of correctly labelled pixels. MIoU, defined as
\begin{equation}
    \text{MIoU} = \frac{1}{C}\sum_{c=1}^C\frac{\sum_{n=1}^N\delta(y_{n,c}, 1)\delta(y_{n,c},\hat{y}_{n,c})}{\sum_{n=1}^N\max(1, \delta(y_{n,c}, 1) + \delta(\hat{y}_{n,c}, 1))}
\end{equation}
is an alternative metric that takes into account class imbalance \cite{csurka2013good} in an image through normalization and is thus a more robust metric compared to accuracy.\\

We train two neural networks on synthetic events generated from video, one using the event representation in \cite{Alonso19cvprw} and one using the EST \cite{Gehrig19iccv}. We evaluate these networks on the test set, and vary the size of the window of events between 10, 50 and 250 ms which was also done in \cite{Alonso19cvprw}. The results from this experiment are summarized in Tab. \ref{tab:seg_test_performance}. We compare against the state of the art method in \cite{Alonso19cvprw}, represented in the last row.
Tab. \ref{tab:seg_test_performance} indicates that the overall accuracy (on 50 ms) for both representations remains within $4\%$ of the $89.8\%$ correctly classified pixels. The difference on the MIoU metric is slightly larger with $45.5\%$ for EST and $48.2\%$ for Alonso et al.'s representation compared to $54.8\%$ if trained on real events. These results indicate that training only on synthetic events yields good generalization to real events, though slightly lower than training on real event data directly. In the next step we want to quantify the gain in when we fine tune on real data.

In a next step we fine tune these models on real data. We do this with a lower learning rate of $10^-4$, and after only two epochs of training we observe large improvements leading to state of the art performance, as captured in Tab. \ref{tab:seg_test_performance}. In fact, our method outperforms existing approaches consistently by an average of 1.2\%. In addition, we see that our method remains moderately robust even with large variations on the event window size.

\begin{table*}[t]
    \centering
    \footnotesize
    \resizebox{0.9\textwidth}{!}{  
    \begin{tabular}{ll|ll|lllll}
         Representation & Fine tuned & Acc. [50 ms] & MIoU [50 ms] & Acc. [10 ms] & MIoU [10 ms] & Acc. [250 ms] & MIoU [250 ms]\\\midrule
         Alonso et al. \cite{Alonso19cvprw} && 86.03& 48.16& 77.25 & 31.76 & 84.24 & 40.18 \\
         EST \cite{Gehrig19iccv} & & 85.93 & 45.48 & 81.11 & 30.70 & 84.49 & 40.66\\
         Alonso et al. & \cmark & 89.36&55.17 &86.06 &39.93 &87.20 &47.85 \\
         EST & \cmark & \textbf{90.19} & \textbf{56.01} & \textbf{87.20} & 45.82 & \textbf{88.64} & \textbf{51.61}\\\midrule
         Alonso et al. & trained on real & 89.76 & 54.81 & 86.46 & \textbf{45.85} & 87.72 & 47.56\\\bottomrule
    \end{tabular}
    }
    \caption{Semantic segmentation performance of different input representations on the test split of \cite{Alonso19cvprw}. \emph{Accuracy} and \emph{MIoU} (\emph{Mean Intersection over Union}). The models are trained on representations of 50 milliseconds (ms) of events and evaluated with a representations of 10 ms and 250 ms of events. The results reported in the last row are taken from \cite{Alonso19cvprw}. This model was trained directly on the real events.}
    \label{tab:seg_test_performance}
    \vspace{-2ex}
\end{table*}
\vspace{-1ex}
\subsubsection{Edge-Cases}
In previous sections we have demonstrated that event datasets generated using our method generalize well to real data and networks trained on these datasets can be fine tuned on real data to enhance performance above state of the art. In this section we investigate how networks trained on synthetic events alone generalize to scenarios in which traditional frames corrupted due to motion blur or over- and under-exposure. In this experiment we use the model trained with EST inputs from synthetic events. Fig. \ref{fig:seg_edge_cases} illustrates two edge cases where frame-based segmentation fails, due to over-exposure (top row) and low contrast (bottom row). We see that in the first case the segmentation network only predicts the background class (top right) since the image is overexposed. In the second case the frame-based segmentation wrongly classifies a person as vegetation which is due to the low contrast in the lower left part of the image. The network using events handles both cases gracefully thanks to the high contrast sensitivity of the event camera. It is important to note that the network never saw real events during training, yet generalizes to edge-case scenarios. This shows that networks trained on synthetic events can generalize beyond the data they were trained with, and do this by inheriting the outstanding properties of events.

\begin{figure}[t]
    \centering
    \includegraphics[width=\linewidth]{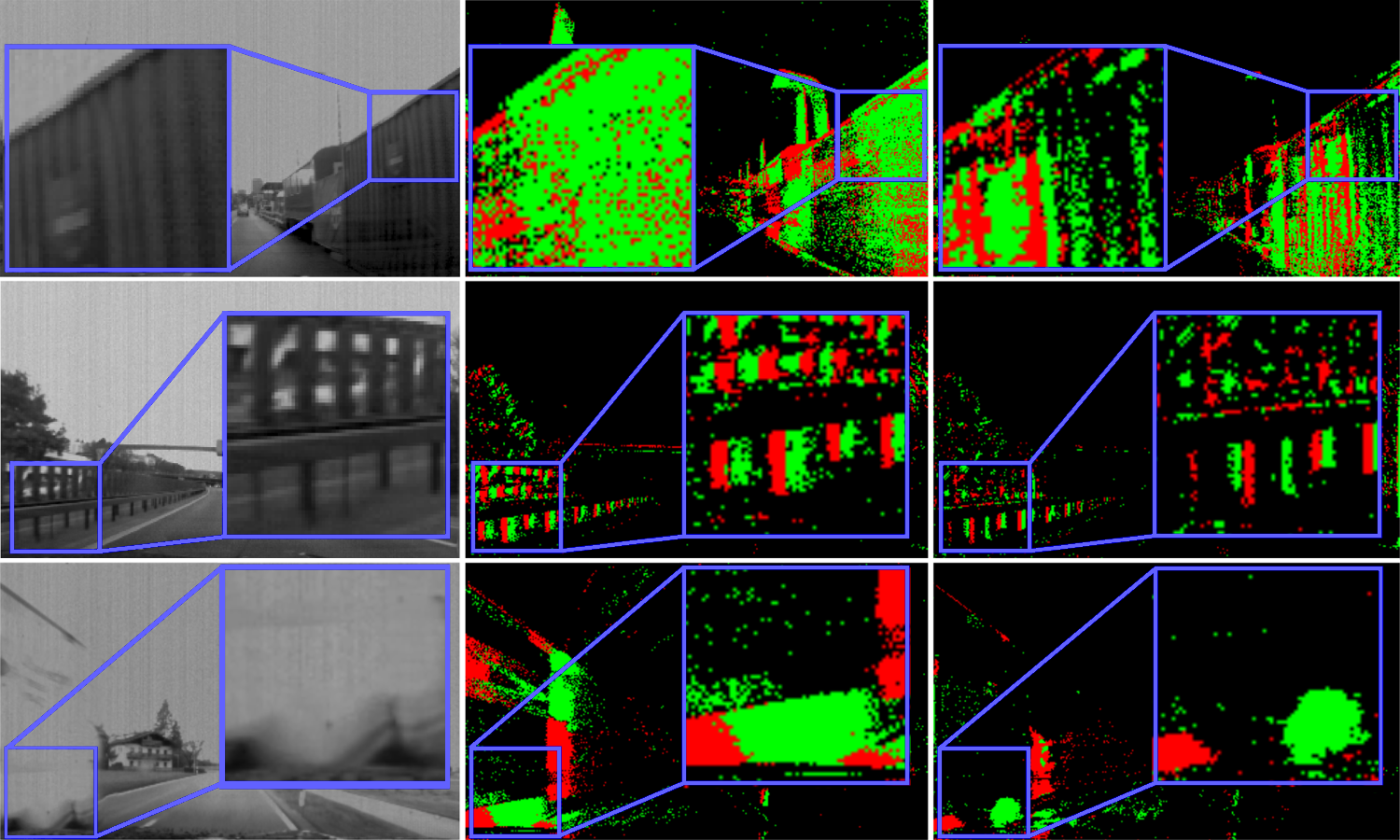}
    \caption{Interpolation artefacts and events in three scenarios (rows). Interpolated frames (left column), real events (middle) and synthetic events (right).
    Positive and negative events colored in green and red respectively. \emph{Row 1 \& 2}: At high optic flows repetitive structures are copied instead of interpolated, leading to missing/wrong events. \emph{Row 3}:  Car tire is incorrectly interpolated (collapses to linear interpolation) due to large optical flow.
    \emph{Best viewed in PDF format}.}
    \label{fig:failure_cases}
    \vspace{-3ex}
\end{figure}

\section{Known Limitations}
One apparent shortcoming is the occurrence of blurry frames in a video dataset. Blurr typically persists in interpolated frames and thus yields suboptimal results when used in combination with the generative model. Furthermore, the generative model does not account for noise that exists in real event cameras. We consider noise modelling an interesting direction of future work that could be incorporated into the proposed framework. Finally, this work builds on frame interpolation methods. While they also have limitations, see Fig. \ref{fig:failure_cases}, it is an active area of research. Consequently, the proposed method can directly benefit from future improvements in frame interpolation techniques.
 
\section{Conclusion}
Over the years, the computer vision community has collected a large number of extensive video datasets for benchmarking novel algorithms. This stands in contrast to the relatively few datasets available to researchers on event-based vision. This work offers a simple, yet effective solution to this problem by proposing a method for converting video datasets into event datasets. The availability of these new synthetic dataset offers the prospect of exploring previously untouched research fields for event-based vision.

The proposed method utilizes a combination of neural network based frame interpolation and widely used generative model for events. We highlight the generalization capability of models trained with synthetic events in scenarios where only real events are available. On top of that, we show that finetuning models (trained with synthetic events) with real events consistently improves results in both object recognition and semantic segmentation.

\section{Acknowledgements}
This work was supported by the Swiss National Center of Competence Research Robotics (NCCR), through the Swiss National Science Foundation, and the SNSF-ERC starting grant.
\section{Appendix} \label{sec:appendix}
\begin{figure*}[h!]
    \centering
    \begin{tabular}{cccc}
        \includegraphics[width=0.2\linewidth]{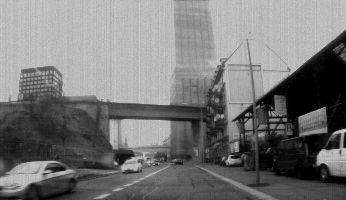} &
        \includegraphics[width=0.2\linewidth]{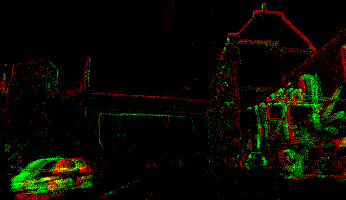} &
        \includegraphics[width=0.2\linewidth]{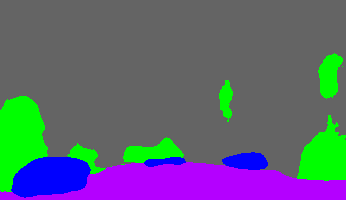} &
        \includegraphics[width=0.2\linewidth]{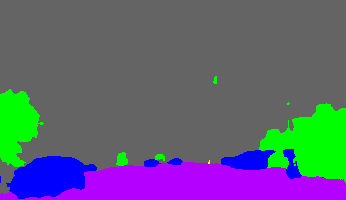}\\
        \includegraphics[width=0.2\linewidth]{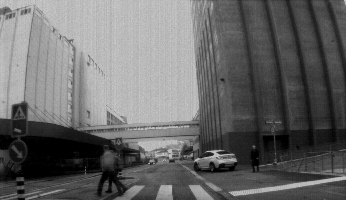} &
        \includegraphics[width=0.2\linewidth]{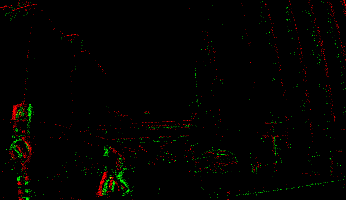} &
        \includegraphics[width=0.2\linewidth]{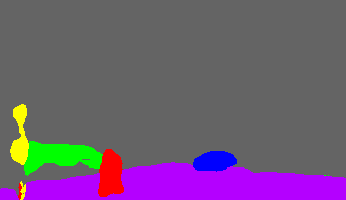} &
        \includegraphics[width=0.2\linewidth]{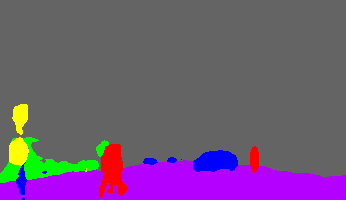}\\
        \includegraphics[width=0.2\linewidth]{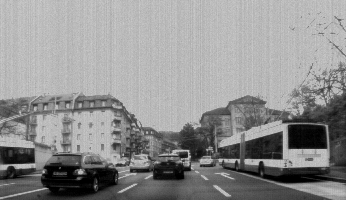} &
        \includegraphics[width=0.2\linewidth]{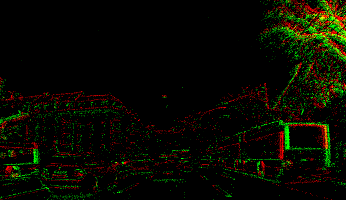} &
        \includegraphics[width=0.2\linewidth]{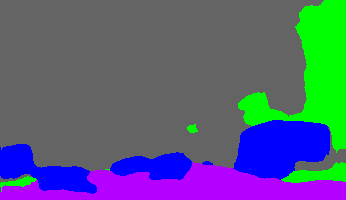} &
        \includegraphics[width=0.2\linewidth]{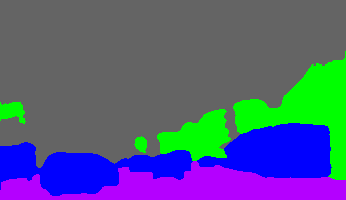}\\
        \includegraphics[width=0.2\linewidth]{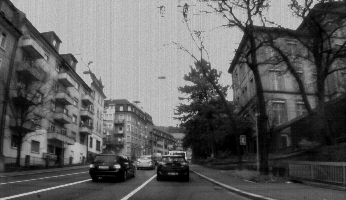} &
        \includegraphics[width=0.2\linewidth]{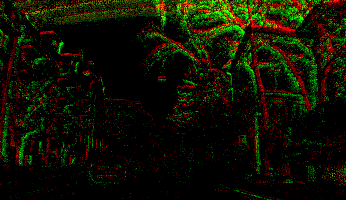} &
        \includegraphics[width=0.2\linewidth]{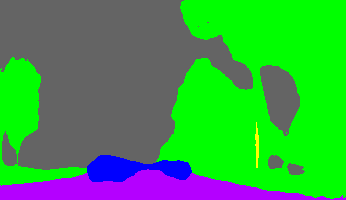} &
        \includegraphics[width=0.2\linewidth]{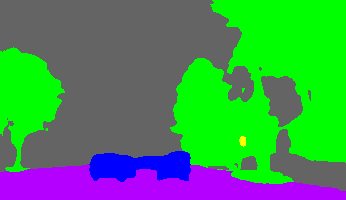}\\
        \includegraphics[width=0.2\linewidth]{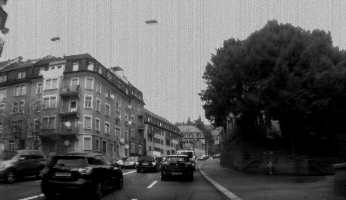} &
        \includegraphics[width=0.2\linewidth]{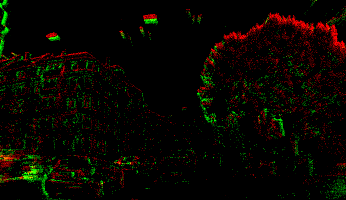} &
        \includegraphics[width=0.2\linewidth]{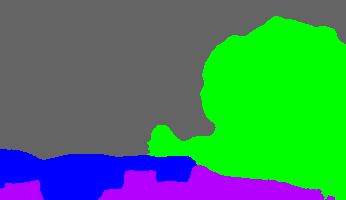} &
        \includegraphics[width=0.2\linewidth]{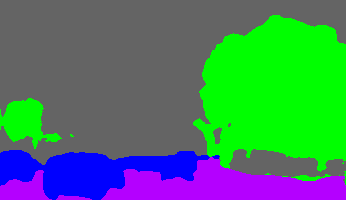}\\
        \includegraphics[width=0.2\linewidth]{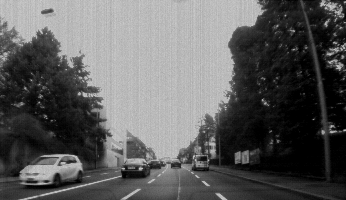} &
        \includegraphics[width=0.2\linewidth]{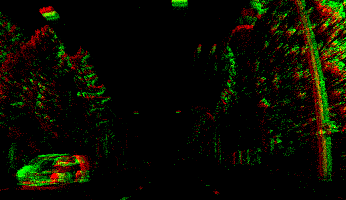} &
        \includegraphics[width=0.2\linewidth]{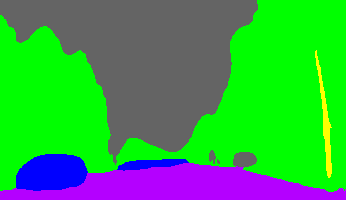} &
        \includegraphics[width=0.2\linewidth]{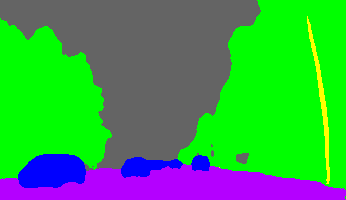}\\
        (a) DAVIS frame & (b) events & (c) prediction & (d) labels
    \end{tabular}
    \caption{A side-by-side view of event-based semantic segmentation predictions for the test set of the DAVIS Driving Dataset (DDD17)   \cite{Binas17icml}. (a) DAVIS frame, (b) visualization of the events used to generate the predictions in (c) and (d) the ground truth labels generated from the frames in (a) \cite{Alonso19cvprw}. The colors represent the following classes: \textcolor{black}{violet: street}; \textcolor{black}{green: vegetation}; \textcolor{black}{red: person}; \textcolor{black}{blue: car}; \textcolor{black}{yellow: object}; \textcolor{black}{gray: background}).}
    \label{fig:app:segmentation_qualitative}
\end{figure*}

In this section we provide additional qualitative results from our segmentation experiment in Sec. 4.2. Several examples from the test set can be viewed in Fig. \ref{fig:app:segmentation_qualitative}. We used the network trained with the EST event representation and fine-tuned on real data. Fine-tuning was only performed for two epochs.\\
%
%

{\small
\bibliographystyle{ieee_fullname}
\bibliography{main.bbl}
}

\end{document}